\documentclass{article} %
\usepackage[preprint]{colm2026_conference}

\usepackage{amsmath}
\usepackage{microtype}
\usepackage{hyperref}
\usepackage{url}
\usepackage{booktabs}
\usepackage{multirow}
\usepackage{colortbl}
\usepackage{xcolor}
\usepackage{graphicx}
\usepackage{tcolorbox}
\tcbuselibrary{breakable, skins}

\usepackage{lineno}

\definecolor{darkblue}{rgb}{0, 0, 0.5}
\hypersetup{colorlinks=true, citecolor=darkblue, linkcolor=darkblue, urlcolor=darkblue}

\title{DeReason: A
Difficulty-Aware Curriculum Improves Decoupled SFT-then-RL Training for
General Reasoning}

\author{%
\textbf{Hanxu Hu}$^{1}$, 
\textbf{Yuxuan Wang}$^{1}$, 
\textbf{Maggie Huan}$^{2}$, 
\textbf{Jannis Vamvas}$^{1}$, 
\textbf{Yinya Huang}$^{3}$,  \\
\textbf{Zhijiang Guo}$^{4,5}$ 
\textbf{and} \textbf{Rico Sennrich}$^{1}$ \\
$^1$University of Zurich \quad $^2$University of Pennsylvania \quad $^3$ETH Zurich \\ $^4$HKUST (GZ) \quad $^5$HKUST \\
}

\begin{document}

\ifcolmsubmission
\linenumbers
\fi

\maketitle

\begin{abstract}
Reinforcement learning with Verifiable Rewards (RLVR) has emerged as a powerful paradigm for eliciting reasoning capabilities in large language models, particularly in mathematics and coding. While recent efforts have extended this paradigm to broader general scientific (STEM) domains, the complex interplay between supervised fine-tuning (SFT) and RL in these contexts remains underexplored. In this paper, we conduct controlled experiments revealing a critical challenge: for general STEM domains, RL applied directly to base models is highly sample-inefficient and is consistently surpassed by supervised fine-tuning (SFT) on moderate-quality responses. Yet sequential SFT followed by RL can further improve performance, suggesting that the two stages play complementary roles, and that how training data is allocated between them matters. Therefore, we propose DeReason, a difficulty-based data decoupling strategy for general reasoning. DeReason partitions training data by reasoning intensity estimated via LLM-based scoring into reasoning-intensive and non-reasoning-intensive subsets. It allocates broad-coverage, non-reasoning-intensive problems to SFT to establish foundational domain knowledge, and reserves a focused subset of difficult problems for RL to cultivate complex reasoning. We demonstrate that this principled decoupling yields better performance than randomly splitting the data for sequential SFT and RL. Extensive experiments on general STEM and mathematical benchmarks demonstrate that our decoupled curriculum training significantly outperforms SFT-only, RL-only, and random-split baselines. Our work provides a systematic study of the interplay between SFT and RL for general reasoning, offering a highly effective and generalized post-training recipe.

\end{abstract}

\section{Introduction}

Reinforcement Learning via Verifiable Rewards (RLVR) has emerged as a powerful paradigm for eliciting reasoning capabilities in large language models (LLMs). Its efficacy has been most thoroughly established in domains with clear outcome-based verification signals, such as mathematics and code reasoning. Recent breakthroughs—exemplified by OpenAI's o1 series and works like DeepSeek-R1 \citep{deepseekr1}, TuluV3 \citep{tulu3}, and TinyZeRO \citep{tinyzero} demonstrate that applying RLVR directly to base models can unlock sophisticated chain-of-thought reasoning. By utilizing rule-based verifiable rewards for math problems and execution-based feedback for competitive programming benchmarks, this approach provides a highly effective training signal for developing systematic reasoning. Consequently, these models exhibit emergent reasoning patterns, including self-verification and reflection. These remarkable successes have sparked considerable interest in understanding exactly how RL transforms model behavior and, crucially, whether such techniques can successfully generalize beyond these narrowly defined verifiable domains.

While RLVR has demonstrated remarkable performance in certain settings, particularly in tasks of math and code, prior work has established that a combination of SFT and RL remains essential for many base models and training scenarios. DeepSeek-R1 itself employs a cold-start SFT stage before RL; Tulu V3 and other production pipelines similarly adopt sequential SFT-then-RL training. However, recent efforts extending RLVR to broader STEM domain, such as General Reasoner \citep{generalreasoner}, WebscaleRL \citep{webscalerl}, have predominantly focused on pure RL approaches, leaving the role of SFT in these settings underexplored. Intuitively, the SFT-then-RL should be even more critical for general STEM reasoning, as acquiring broader domain knowledge. This naturally raises the question: given that both SFT and RL play complementary roles, how should training data be allocated between the two stages in general domains? We investigate this through a difficulty-aware curriculum that partitions data to match each stage's training. Importantly, our approach operates at the data selection level rather than proposing algorithmic modifications for combining SFT and RL training. This makes it orthogonal to existing algorithmic improvements \citep{huang2025blendingsupervisedreinforcementfinetuning, yan2025learningreasonoffpolicyguidance} and can be directly used in various training frameworks and toolkits.

Concretely, we first separately train models with pure SFT and pure RLVR on general STEM reasoning tasks and compare their resulting capabilities by controlling amount of training data. Our findings reveal a clear division of labor: across both mathematical and broader STEM domains, pure RLVR applied directly to a base model is consistently and significantly outperformed by SFT. Based on this observation, we propose \textbf{DeReason}, a difficulty-based decoupled training strategy. Specifically, we introduce \textit{reasoning intensity} as the partitioning criterion and employ an LLM to score each training instance on a scale of 1 to 5. Problems that primarily require knowledge recall or straightforward application of known facts receive low reasoning intensity scores, while problems demanding multi-step derivation and reasoning receive high scores. We then allocate low reasoning intensity data to SFT—as these knowledge-recall-oriented problems are precisely where distillation from a stronger teacher is most efficient—and reserve high reasoning intensity data for RLVR, where the model benefits from exploring complex reasoning paths beyond the teacher's demonstrations.

Our contributions are summarized as follows:

1) We systematically analyze the interplay of SFT and RLVR across both math and general STEM tasks, demonstrating that for small models, SFT serves as an indispensable distillation and cold-start mechanism that vastly outperforms pure RLVR.

2) \textsc{DeReason} Curriculum: We propose a novel, decoupled training strategy, demonstrating that partitioning data by difficulty: SFT on easy/broad data followed by RLVR on selected hard data, significantly outperforms pure SFT, pure RLVR, or random SFT-then-RLVR baselines.

3) Detailed Behavioral Analysis: We provide a fine-grained analysis of the training dynamics. Specifically, we evaluate the impact of different difficulty selection distributions and characterize how SFT and RLVR uniquely shape model behavior, detailing their distinct effects on policy entropy, response length evolution, and reward optimization.

\section{Motivation}
\label{sec:motivation}

Recent work has demonstrated the remarkable effectiveness of reinforcement learning with verifiable rewards (RLVR) for improving reasoning capabilities of large language models, particularly in mathematical domains~\citep{tinyzero, drgrpo, deepseekr1}. These successes have led to a growing consensus that RL-based post-training is broadly superior to supervised fine-tuning (SFT) for eliciting reasoning abilities. However, we argue that this conclusion may be premature and warrants more careful examination under controlled experimental conditions.

\begin{figure}[t]
    \centering
    \includegraphics[width=0.85\linewidth]{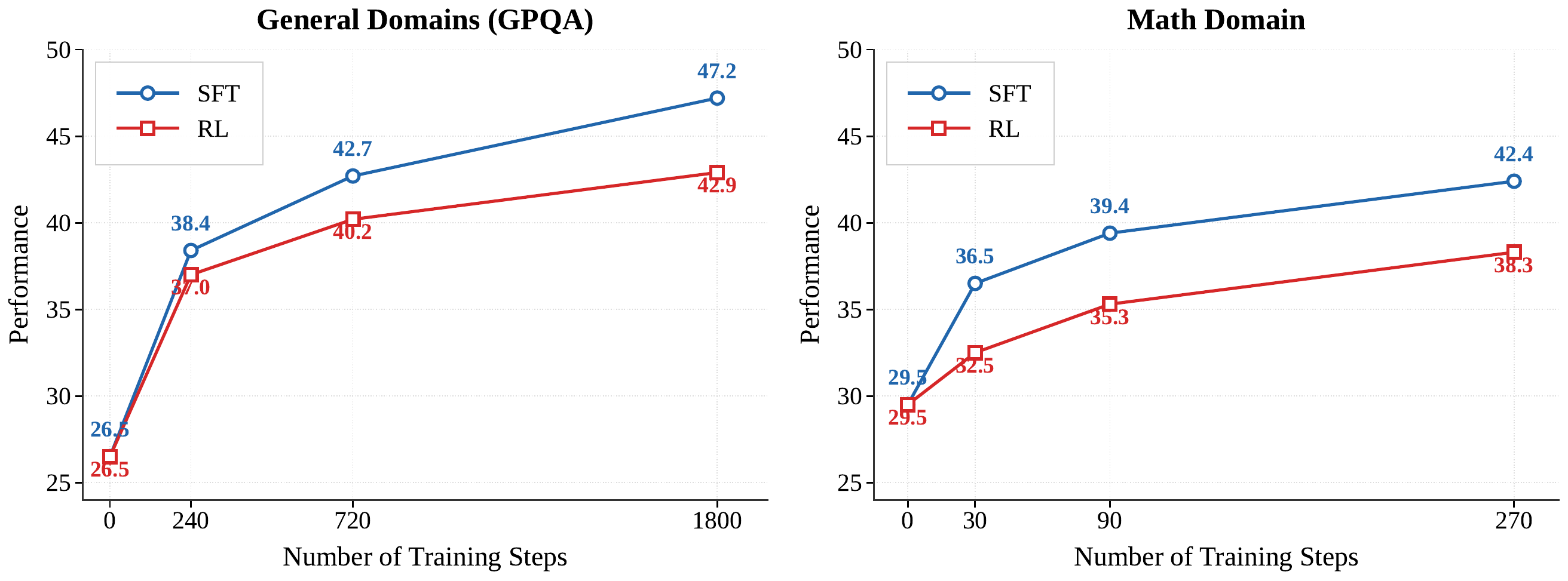}
    \caption{
        Scaling behavior of RL and SFT on both general STEM and math domains, trained on the same set of problems. In the general domain, we report pass@1 averaged across 8 runs on GPQA-Diamond. In the math domain, we report the same metric averaged across AIME24, AIME25, and MATH500. In both settings, SFT with moderate-quality model responses outperforms RL as training data increases.
    }
    \label{fig:scaling}
\end{figure}

\paragraph{RL vs.\ SFT under Controlled Comparison.}
To rigorously assess the relative merits of RL and SFT, we conduct a series of controlled experiments where both methods are trained on \emph{exactly the same set of problems}, varying only the amount of training data. For SFT, we use responses generated by a moderate-capability model rather than a frontier model, ensuring that the supervision signal is not too strong. As shown in Figure~\ref{fig:scaling}, we evaluate on both general STEM reasoning (GPQA-Diamond, averaged across 8 runs at pass@1) and mathematical reasoning (pass@1 averaged across AIME24, AIME25, and MATH500).

Our results reveal that \emph{SFT consistently outperforms RL as training data scales up in both domains}. In the math domain, SFT with moderate-quality responses already achieves competitive or superior performance compared to RL trained on the same problems. In general STEM domains, the gap is similar. RL struggles to match SFT performance even with increasing data, suggesting that outcome-based reinforcement alone is insufficient for acquiring the broad domain knowledge required for general scientific reasoning.

We attribute the advantage of SFT to its superior \emph{sample efficiency}: direct imitation of moderate-quality solutions provides a stronger learning signal than outcome-based reinforcement, where the model must discover effective reasoning paths through noisy exploration—particularly challenging for a small base model without prior fine-tuning. Moreover, both mathematical and general STEM reasoning require \emph{domain knowledge} (e.g., physics formulae, algebraic identities) that is difficult to acquire through trial-and-error alone, whereas SFT offers a more direct pathway for knowledge consolidation. These observations suggest that SFT and RL have complementary strengths: SFT excels at efficient knowledge acquisition, while RL can push performance beyond the supervision signal on sufficiently challenging problems. This motivates \textsc{Dereason}, a \emph{difficulty-based data decoupling} strategy that allocates easier samples to SFT for knowledge and skill acquisition, and reserves difficult samples for RL to push the reasoning frontier beyond what imitation alone can achieve.

\subsection{Preliminaries}

\paragraph{Supervised Fine-Tuning (SFT).}
Given a dataset $\mathcal{D}_{\text{SFT}} = \{(x_i, y_i)\}_{i=1}^{N}$ of problem-response pairs, SFT optimizes the policy $\pi_\theta$ by maximizing the log-likelihood of reference responses.

\paragraph{Reinforcement Learning with GRPO.}
Group Relative Policy Optimization (GRPO)~\citep{shao2024deepseekmath} estimates advantages from group-level statistics, eliminating the need for a separate value model. For each prompt $x$, GRPO samples a group of $G$ responses $\{o_1, \ldots, o_G\}$ from the current policy $\pi_\theta$, each receiving a reward $r_i$ from a reward function $R(\cdot)$. Advantages are computed by normalizing rewards within each group. The GRPO objective is:
\begin{equation}
    \mathcal{L}_{\text{GRPO}}(\theta) = -\mathbb{E}_{x \sim \mathcal{D}_{\text{RL}}} \left[ \frac{1}{G} \sum_{i=1}^{G} \min\left( \rho_i \hat{A}_i, \; \text{clip}(\rho_i, 1{-}\varepsilon, 1{+}\varepsilon) \hat{A}_i \right) - \beta \, D_{\text{KL}}\left(\pi_\theta \| \pi_{\text{ref}}\right) \right],
\end{equation}
where $\rho_i = \frac{\pi_\theta(o_i \mid x)}{\pi_{\text{ref}}(o_i \mid x)}$ is the importance sampling ratio, $\hat{A}_i$ is the group-normalized advantage, $\varepsilon$ is the clipping parameter, and $\beta$ controls KL regularization strength.

\paragraph{Verification for General Reasoning}
Reasoning tasks in mathematics and code often admit deterministic reward signals, e.g., by matching numerical answers or executing test cases. However, for general scientific domains, answers frequently involve free-form explanations or qualitative reasoning that cannot be assessed by rule-based checkers. We therefore follow \cite{generalreasoner} to adopt a \emph{model-based verifier} to judge response correctness:
\begin{equation}
    R(x, o) = \begin{cases} 1 & \text{if } \mathcal{V}_\theta\bigl(\textsc{Extract}(o),\; a^*,\; x\bigr) = \text{True}, \\ 0 & \text{otherwise}, \end{cases}
\end{equation}
where $a^*$ is the ground-truth answer for prompt $x$, $\textsc{Extract}(\cdot)$ extracts the final answer from model response $o$, and $\mathcal{V}_\theta$ is a language-model-based verifier that assesses semantic equivalence between the extracted answer and $a^*$ conditioned on the question $x$. Unlike rule-based verification, the model-based verifier can handle diverse answer formats in scientific reasoning, including qualitative explanations, approximate numerical values, and multi-part derivations.

\subsection{DeReason: Difficulty-Based Data Decoupling}
\paragraph{Overall Pipeline.}
Let $\mathcal{D} = \{(x_i, a_i^*)\}_{i=1}^{N}$ denote the full training set of problems with ground-truth answers. Our method proceeds in three stages:

\begin{enumerate}
    \item \textbf{Difficulty Estimation}: Assign a difficulty score $d_i \in [1, 5]$ to each problem $x_i$ (described below).
    \item \textbf{Data Partitioning}: Based on the difficulty scores, partition $\mathcal{D}$ into an SFT subset $\mathcal{D}_{\text{SFT}}$ (easier, broader) and an RL subset $\mathcal{D}_{\text{RL}}$ (harder, focused):
    \begin{equation}
        \mathcal{D}_{\text{SFT}} = \{(x_i, a_i^*) \in \mathcal{D} \mid d_i \leq \tau \}, \quad \mathcal{D}_{\text{RL}} = \{(x_i, a_i^*) \in \mathcal{D} \mid d_i > \tau \},
    \end{equation}
    where $\tau$ is a difficulty threshold. For $\mathcal{D}_{\text{SFT}}$, we generate reference responses $y_i$ using a moderate teacher model (e.g., Qwen3-4B-Instruct) to construct SFT pairs.
    \item \textbf{Curriculum Training}: First perform SFT on $\mathcal{D}_{\text{SFT}}$ to obtain $\pi_{\text{SFT}}$, then apply GRPO on $\mathcal{D}_{\text{RL}}$ initialized from $\pi_{\text{SFT}}$.
\end{enumerate}

\paragraph{Difficulty Estimation.}
We employ an LLM to estimate problem difficulty. To avoid reliance on external proprietary models, we intentionally use an instruct model of the same size (Qwen3-4B-Instruct here) to the policy model as the judge.

\subparagraph{LLM-based Scoring.}
We prompt the same size instruct LLM to directly assess the difficulty of each problem on a scale from 1 to 5, considering factors such as the number of reasoning steps, prerequisite domain knowledge, and potential for error. The detailed prompt $p_{\text{diff}}$is shown in Appendix \ref{app:reasoning_complexity_prompt} and it is used for instructing the model to output a difficulty score $s_i \in \{1, 2, 3, 4, 5\}$. We then assign problems with high difficulty scores ($s_i \geq 4$) to the RL training set.

\section{Experiments}
\subsection{Training}
We use Qwen3-4B-Base as base model for both SFT and RL. In SFT experiments, we use batch size as 128 and learning rate as 1e-5 under Llama-Factory framework \citep{zheng-etal-2024-llamafactory}. In RL, we use VeRL \citep{sheng2024hybridflow} for all experiments, and set max response length as 8192, training batch size as 128, and mini batch as 64, the learning rate is set as 1e-6. We use Qwen3-4B-Instruct-2507 to get all responses for SFT data, as it is only a small size instruct model, which makes us not depend too much on the cabability of external strong model in making SFT data. For validating the generalization of our method, we conduct experiments on two different datasets, WebInstruct-Verified and Webscale-RL, both of them focus on STEM domains. 
\subsection{Evaluation}
To evaluate the general reasoning of the model comprehensively, we use multiple challenging general reasoning datasets: 

MMLU-Pro \citep{wang2024mmlupro}: An enhanced version of MMLU that increases answer choices from 4 to 10 and incorporates more reasoning-intensive questions, reducing the chance of guessing correctly by memorization alone and providing better discrimination between strong models.

GPQA-Diamond \citep{rein2024gpqa}: A multiple-choice benchmark designed to require genuine expert-level knowledge, with questions authored by PhD-level domain experts such that non-experts score near random chance (~34 \%). The Diamond subset represents the highest-quality, most rigorously filtered portion of the full dataset.

SuperGPQA \citep{du2025supergpqa}: A large-scale extension of GPQA covering 285 disciplines with tens of thousands of graduate-level questions, targeting long-tail subject knowledge and addressing the limited disciplinary coverage of existing benchmarks.

BBEH \citep{bbeh}: BIG-Bench Extra Hard, an upgraded successor to BIG-Bench Hard that redesigns existing tasks to remain challenging for state-of-the-art models, focusing on complex multi-type reasoning (logical, spatial, arithmetic, etc.) with the explicit goal of maintaining a significant gap between model and human performance.

\subsection{Baselines}
We compare with various baselines in previous works. Specifically, we use the training data of WebInstruct-Verified \citep{generalreasoner} and WebScaleRL \citep{webscalerl}, and conduct GRPO on them from base model following their training settings, and use the verifier model from \citet{generalreasoner} to provide reward scores. 

\subsection{Main results}

For we validate our method in two datasets, Webinstruct-Verified and Webscale-RL, we report results of models training on both of them separately. 

We compared with previous baselines, more specifically, we mainly compare the results of only SFT and only RL on WebscaleRL and Webinstruct-Verified respectively. Besides, we also used the same data, first training on easy data, then training on selected difficult data, performing this training using only SFT and only RL for further ablation. The selected RL here means we used an LLM for selecting those problems scored as 4 and 5 for difficulty. It shows that SFT-only can be clearly better than RL-only when testing wit the same training data in all benchmarks. Using our pipeline, SFT on easy subset and RL on hard subset can further boost the performance, leading to best results in 4B models. At the same time, our model also outperforms all previous models and baselines in similar scale. Additionally, we also observe that, on easy benchmarks like MMLU-Pro, the gap between our approach and SFT-only baselines is small, or our approach achieves even worse results than the SFT-only baseline, but on hard benchmarks like BBEH, which require more reasoning than knowledge retrieval, our pipeline yields a clear improvement compared to other baselines.

\begin{table*}[ht]
\centering
\caption{Main results on reasoning benchmarks. Bold indicates best performance within each model group.}
\label{tab:main}
\resizebox{\textwidth}{!}{%
\begin{tabular}{lcccccc}
\toprule
\textbf{Models} & \textbf{MMLU-Pro} & \textbf{GPQA-D} & \textbf{SuperGPQA} & \textbf{BBEH} & \textbf{AVG} \\
\midrule
\multicolumn{6}{l}{\textit{Baselines}} \\
\midrule
 GPT-4o                       & 74.6          & 50.0          & 46.3          & 22.3          & 48.3 \\
 QwQ-32B                      & 52.0          & 54.5          & 43.6          & 22.6          & 43.2 \\
DeepSeek-R1                  & 84.0          & 71.5          & 59.9          & 34.9          & 62.6 \\
Qwen2.5-7B-Base              & 47.7          & 29.3          & 26.7          & 8.0           & 27.9 \\
Qwen2.5-7B-Instruct          & 57.0          & 33.8          & 30.7          & 12.2          & 33.4 \\
 Open-Reasoner-Zero           & \textbf{59.4} & 36.6          & 32.8          & 12.2          & 35.3 \\
 SimpleRL-Qwen2.5-7B-Zoo      & 51.5          & 24.2          & 29.9          & 11.9          & 29.4 \\
\midrule
\multicolumn{6}{l}{\textit{4B Models (Webinstruct-verified)}} \\
\midrule
Qwen3-4B-Base                          & 51.6   & 26.5   & 25.4   & 8.1   & 27.9 \\
Webinstruct-V (RL only)                    & 62.8 & 42.9 & 32.5 & 12.2 & 37.6 \\
Webinstruct-V (SFT only)                   & 68.6 & 46.8 & 38.4 & 13.5 & 41.8 \\
Webinstruct-V (SFT then selected SFT)      & 68.6 & 45.6 & 38.2 & 13.0 & 41.4 \\
Webinstruct-V (SFT then random RL)         & 68.6   & 47.8   & 39.4   & 15.8  & 42.9 \\
\rowcolor{blue!6}
Webinstruct-V (Ours, SFT easy + RL hard) & 68.4 & \textbf{50.0} & \textbf{40.2} & \textbf{16.7} & \textbf{43.8} \\
\midrule
\multicolumn{6}{l}{\textit{4B Models (Webscale-RL)}} \\
\midrule
Webscale (RL only)                     & 55.4   & 34.0 & 30.9   & 10.1   & 32.6 \\
Webscale (SFT only)                    & 60.7 & 39.2 & 37.3 & 13.4 & 37.7 \\
\rowcolor{blue!6}
Webscale (Ours, SFT easy + RL hard) & 60.3 & \textbf{43.7} & \textbf{38.8} & \textbf{15.7} & \textbf{39.6} \\
\bottomrule
\end{tabular}%
}
\end{table*}

\section{Analysis}
\subsection{Distribution of data in different difficulty}

\begin{figure}[b]
    \centering
    \includegraphics[width=0.8\linewidth]{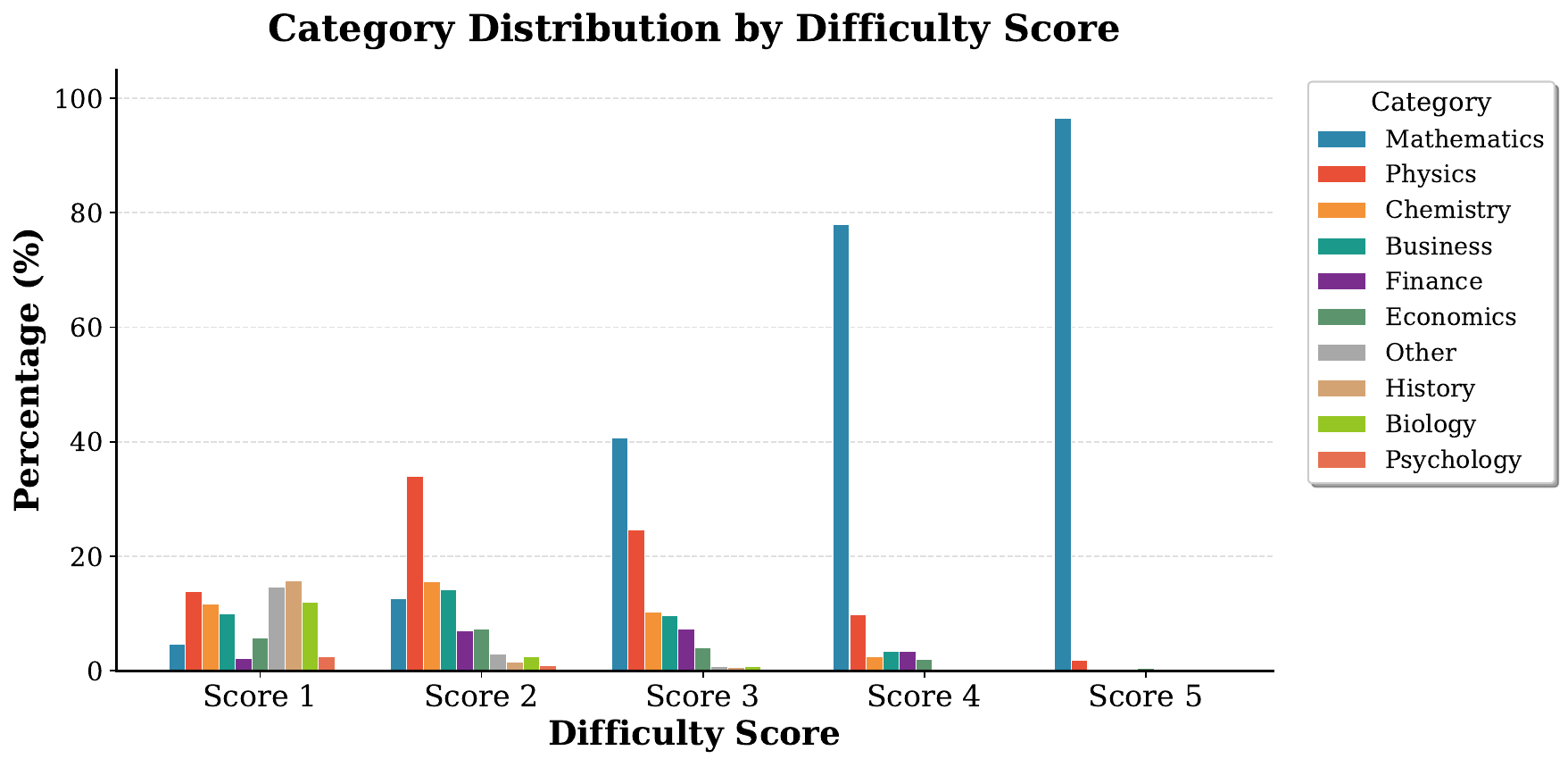}
    \caption{
        The category distribution across difficulty scores.
    }
    \label{fig:distribution}
\end{figure}

Figure \ref{fig:distribution} shows the category distribution of training data of WebInstruct-Verified across difficulty levels, as judged by LLMs on a scale from 1 to 5. At lower difficulty scores, the data is distributed relatively evenly across diverse categories such as History, Biology, Business, and Psychology. As difficulty increases, however, the distribution becomes increasingly concentrated in Mathematics and Physics. At the highest difficulty levels (scores 4 and 5), Mathematics dominates overwhelmingly—comprising roughly 78\% and 96\% of samples, respectively. This trend suggests that easy samples tend to cover broad, knowledge-oriented topics, while harder samples are predominantly reasoning-intensive, consistent with our assumption that difficulty correlates with the shift from knowledge recall to complex reasoning.
\subsection{Performance on mathematic task}
We also evaluate our method with SFT-only and RL-only baselines on mathematic tasks. We found that in most situations our method is better than these baselines in both training data we used, and the performance follow similar trend with STEM general reasoning benchmarks.

\begin{table*}[ht]
\centering
\caption{Math reasoning results on AIME24, AIME25, and MATH500.}
\label{tab:math}
\resizebox{0.7\textwidth}{!}{%
\begin{tabular}{lccc}
\toprule
\textbf{Models} & \textbf{AIME24} & \textbf{AIME25} & \textbf{MATH500} \\
\midrule
\multicolumn{4}{l}{\textit{4B Models (Webinstruct-verified)}} \\
\midrule
WebIns-V (RL only)                          & 20.0 & 15.4 & 80.6 \\
WebIns-V (SFT only)                         & 22.0 & 17.6 & 82.6 \\
\rowcolor{blue!6}
WebIns-V (Ours, SFT then selected RL) & \textbf{22.1} & \textbf{18.0} & \textbf{84.1} \\
\midrule
Webscale (RL only)                           & 21.3 & 14.0 & 81.6 \\
Webscale (SFT only)                          & 26.3 & \textbf{23.3} & 87.5 \\
\rowcolor{blue!6}
Webscale (Ours, SFT then selected RL)      & \textbf{27.7} & 20.7 & \textbf{88.1} \\
\bottomrule
\end{tabular}%
}
\end{table*}

\subsection{Using different difficulty selection for RL}
\begin{figure}[t]
    \centering
    \includegraphics[width=\linewidth]{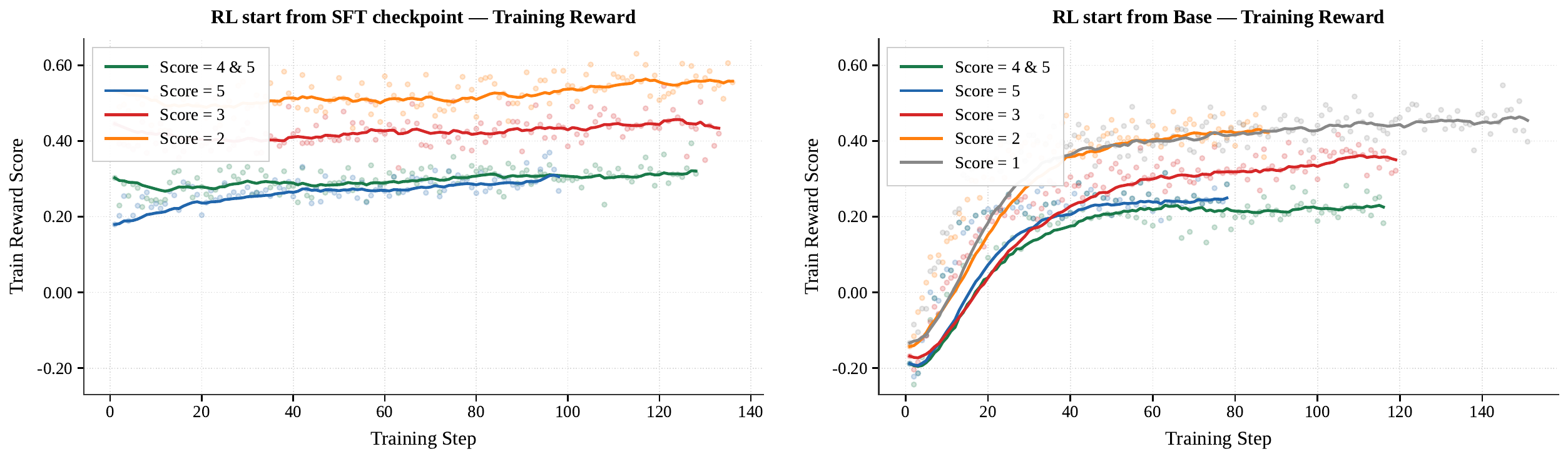}
    \caption{
        Reward score during RL training with data in different difficulty score (from 1 to 5) from Webinstruct-verified. The left sub figure is starting from SFT checkpoint, right sub figure is starting from base checkpoint
    }
    \label{fig:reward_score}
\end{figure}

\begin{figure}[t]
    \centering
    \includegraphics[width=0.9\linewidth]{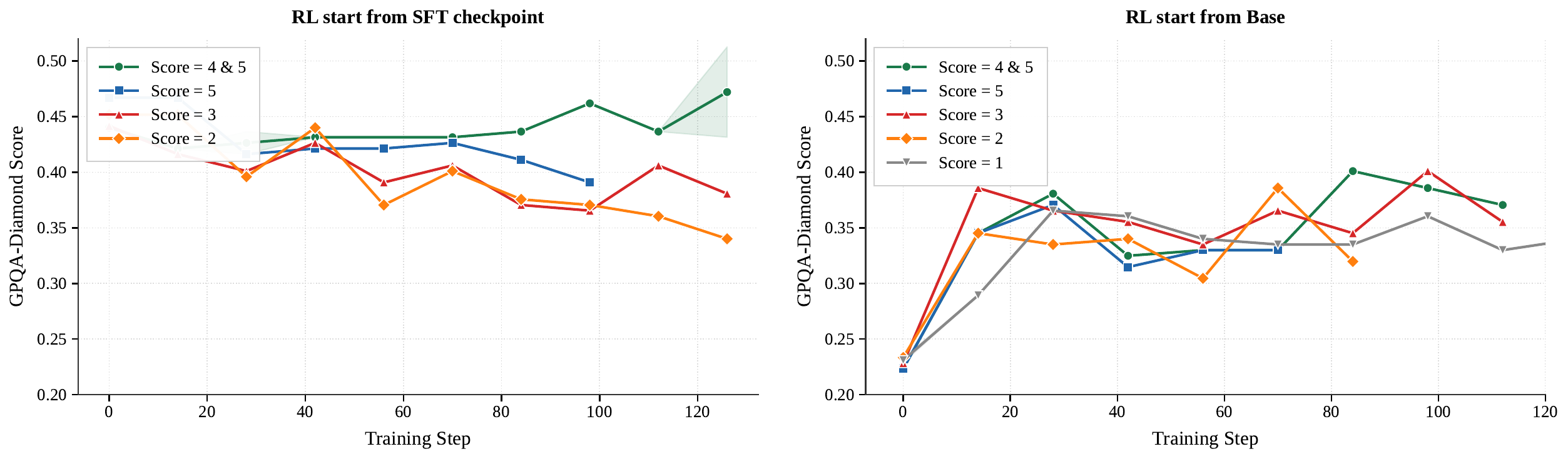}
    \caption{
        GPQA-diamond score during RL training from SFT checkpoint and base model respectively. Pass@1 averaged across 8 runs is computed for each checkpoints.
    }
    \label{fig:val_score}
\end{figure}

We conduct further analysis on using data with different difficulty for RL separately from both base model as starting checkpoint and SFTed model as starting checkpoint, and we show the training reward scores in Figure \ref{fig:reward_score}, it shows that the initial reward of the SFT checkpoint is higher than that of the base model, and there is a slight improvement in subsequent steps; the base model shows a significant improvement in the first 40 steps, but then the performance tends to level off.  While Figure \ref{fig:val_score} shows SFT checkpoint has higher performance, but gradually declined except for the 4 and 5 subset. Base performance was relatively low at first, but then slowly increased.

\subsection{Comparison of response length}
\begin{figure}[t]
    \centering
    \includegraphics[width=\linewidth]{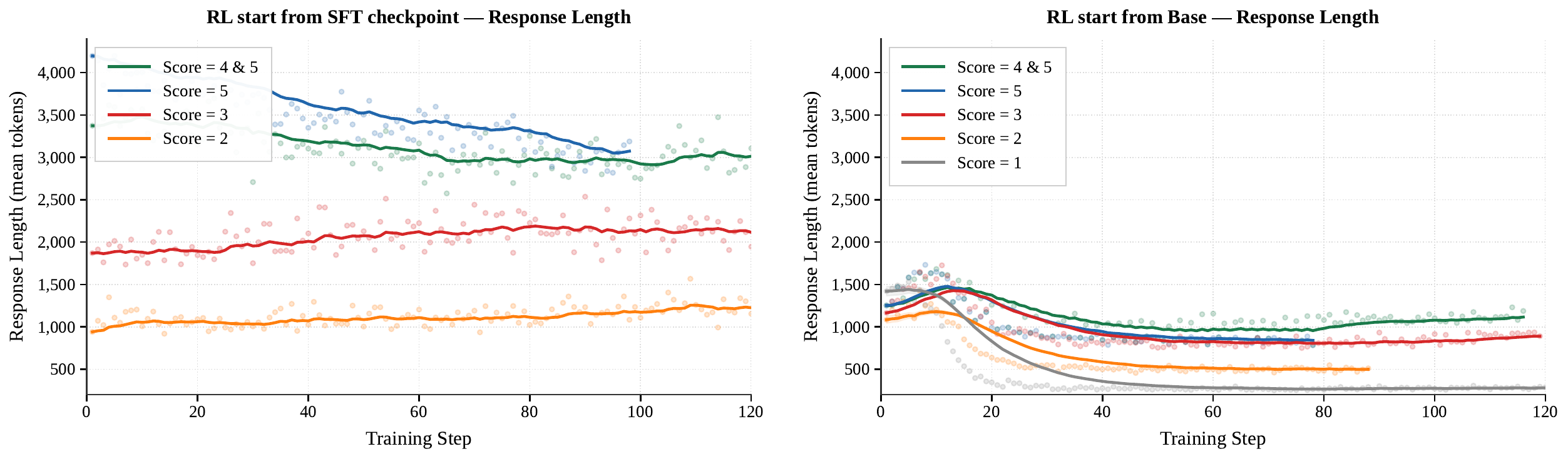}
    \caption{
        Response length during RL training with data in different difficulty scores.
    }
    \label{fig:response_length}
\end{figure}

Figure \ref{fig:response_length} shows the evolution of response length throughout RL training, broken down by verifiable reward score. Starting from the SFT checkpoint (left), the model inherits verbose generation behavior, and RL progressively shortens responses—most notably for high-scoring outputs, which drop from approximately 4,200 to 3,000 tokens. From the base model (right), responses across all score levels initially share similar lengths (~1,200 tokens), but rapidly diverge: responses for high-scoring questions sustain or grow in length, whereas low-scoring ones shrink to below 500 tokens. This score-dependent bifurcation is far more pronounced in the base model setting, where the gap between Score 5 and Score 1 responses widens to over 1,000 tokens within the first 40 steps. In the SFT setting, this gap is less dramatic, as RL primarily acts as a compression mechanism that preserves the existing length–quality hierarchy while uniformly reducing verbosity.

\subsection{Observation of entropy}

\begin{figure}[t]
    \centering
    \includegraphics[width=0.6\linewidth]{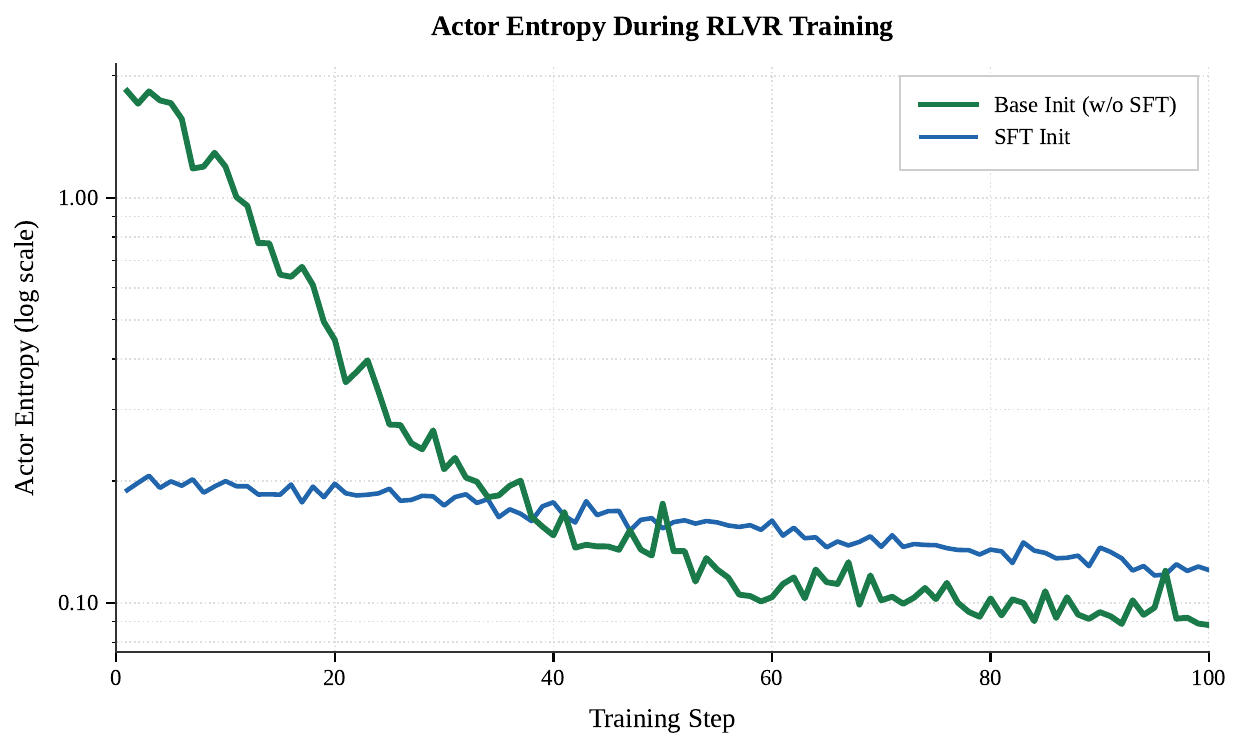}
    \caption{
        Actor's entropy during RL training with different initial checkpoint.
    }
    \label{fig:entropy}
\end{figure}

Figure \ref{fig:entropy} presents the actor entropy (on a log scale) throughout RL training for both initialization settings. The base model begins with substantially higher entropy ($\approx$2.0), reflecting its broad and less constrained output distribution, and undergoes a steep decline during the first 20 steps before gradually stabilizing below 0.10. In contrast, the SFT-initialized model starts with much lower entropy ($\approx$0.30), as supervised fine-tuning has already concentrated the policy distribution, and exhibits a slower, more moderate decrease over the course of training. Notably, the base model's entropy eventually drops below that of the SFT model, suggesting that RL from the base model ultimately converges to a more deterministic policy. This indicates that while SFT pre-narrows the policy space, RL from the base model achieves even sharper specialization through reward-driven exploration and exploitation.

\section{Related work}

Reinforcement learning with verifiable rewards (RLVR)~\citep{tulu3} has driven recent reasoning breakthroughs: GRPO~\citep{shao2024deepseekmath},  o1~\citep{openai_o1} and DeepSeek-R1~\citep{deepseekr1} demonstrated that large-scale RL can incentivize strong reasoning capabilities, also reproduced at small scale by TinyZero~\citep{tinyzero}. Subsequent works refine the training algorithm: Dr.\ GRPO~\citep{drgrpo} and DAPO~\citep{dapo}. The RLVR training paradigm has been widely adopted in production models~\citep{qwen3,kimik2,glm5}.
Additionally, several works explore verifier-free alternatives that use likelihood-based signals as rewards~\citep{verifree,rlpr,darl,likelihoodreward}. Recently, some works  \citep{huang2025blendingsupervisedreinforcementfinetuning, yan2025learningreasonoffpolicyguidance} try to blend SFT and RL algorithmically and achieved better mathematic performance. While RLVR has proven highly effective for mathematical and code reasoning, extending this paradigm to general reasoning remains challenging.

Several recent efforts aim to bridge this gap by scaling verifiable training data across domains~\citep{mammoth2,generalreasoner,webscalerl} and improving RL training with multi-domain, multi-format data~\citep{crossthink,crossingreward}. A systematic study by \citet{guru} further reveals that RL transfer is highly domain-dependent. Yet how training data of varying difficulty should be allocated across post-training stages remains underexplored.

\label{sec:difficulty}

The choice of training data difficulty critically affects RL effectiveness in reasoning. \citet{hardexamples} show that training GRPO on the hardest 10\% of examples that the base model fails most often yields gains of up to 47\%; by contrast, easy examples produce only 3--15\% improvement.
E2H Reasoner~\citep{e2hreasoner} takes a curriculum learning~\citep{bengio2009curriculum} approach, scheduling RL training from easy to hard problems and showing that this ordering improves reasoning over vanilla RL training alone.
Another approach is to adaptively adjust problem difficulty during RL training: DEPO~\citep{depo_zhao} uses an online difficulty estimator to filter trivial or overly complex samples before rollout, \citet{dots} select moderate-difficulty questions via an attention-based estimator, and SEC~\citep{sec} formulates difficulty selection as a multi-armed bandit that adaptively chooses which difficulty category to train on at each step.
\citet{dataefflvr} bridge both paradigms by combining offline curation with online explorability filtering. Despite these advances, all prior works focus on difficulty selection within the RL paradigm, without examining how difficulty should inform the allocation of data across distinct post-training stages.

Along a different axis, several works characterize the distinct roles of SFT and RL in post-training: SFT memorizes training data while RL generalizes to unseen variants \citep{sftmemorizesrlgeneralizes}; mechanistically, SFT expands correct reasoning trajectories whereas RL compresses incorrect ones \citep{rlsqueezessftexpands}; theoretically, their objectives are inherently coupled in parameter space, so the second stage necessarily degrades the first \citep{nondecouplingsfrl}; and under math-only training, RL preserves general-domain representations while SFT induces drift \citep{mathtransfer}. Recent attempts to blend SFT and RL algorithmically have improved mathematical performance \citep{huang2025blendingsupervisedreinforcementfinetuning, yan2025learningreasonoffpolicyguidance}, yet these studies remain confined to mathematical tasks, leaving open how their findings extend to general STEM domains and how data should be allocated to match each stage's learning dynamics. Notably, because DeReason operates purely at the data selection level, it is orthogonal to algorithmic advances in SFT or RL and can be readily integrated into existing training pipelines.

\section{Conclusion}
In this paper, we introduce DeReason, a difficulty-aware curriculum training strategy for general STEM reasoning. Through controlled comparisons of pure SFT and RLVR, we observe SFT consistently outperforms RLVR when applied directly to a base model. Motivated by this finding, we propose partitioning training data by reasoning intensity—routing knowledge-recall problems to SFT and reasoning-heavy problems to RLVR. This simple data-level strategy yields consistent improvements over random allocation baselines in STEM domains.  We hope this work encourages further investigation into principled data allocation strategies for multi-stage LLM post-training.

\section*{Acknowledgments}
HX, JV, and RS acknowledge
funding by the Swiss National Science Foundation (project InvestigaDiff; no. 10000503). HX, RS and YY also acknowledge the compute resource provided by the Swiss AI Initiative.

\bibliography{colm2026_conference}
\bibliographystyle{colm2026_conference}

\appendix
\newpage
\section{Appendix}
\subsection{Reasoning Complexity Rating Prompt}
\label{app:reasoning_complexity_prompt}

Table~\ref{tab:reasoning_complexity_prompt} presents the prompt template used to assess the reasoning complexity of questions on a 1--5 scale.

\begin{table*}[ht]
\centering
\small
\begin{tcolorbox}[
    colback=gray!5!white,
    colframe=gray!75!black,
    title={\textbf{Reasoning Complexity Rating Prompt}},
    fonttitle=\small,
    breakable,
    left=4pt, right=4pt, top=4pt, bottom=4pt
]
\ttfamily\footnotesize
Please rate the reasoning complexity of the given question using the scale below.\\[4pt]
Question: \textless question\textgreater\ \{question\} \textless/question\textgreater\\[6pt]

\textnormal{\textbf{ReasoningRequired Scale:}}\\[4pt]

\textnormal{\textbf{1 - Minimal:}} Direct recall or single-step lookup\\
\hspace*{1em}\textbullet\ Answer is immediately retrievable from memory/knowledge\\
\hspace*{1em}\textbullet\ No logical steps or decomposition needed\\
\hspace*{1em}\textbullet\ Example: ``What is the capital of France?''\\[4pt]

\textnormal{\textbf{2 - Simple:}} One or two logical steps\\
\hspace*{1em}\textbullet\ Requires basic inference or straightforward calculation\\
\hspace*{1em}\textbullet\ Single relationship or comparison\\
\hspace*{1em}\textbullet\ Example: ``If apples cost \$2 and you have \$5, how many can you buy?''\\[4pt]

\textnormal{\textbf{3 - Moderate:}} Multiple steps with clear structure\\
\hspace*{1em}\textbullet\ Requires 3--5 logical steps or simple decomposition\\
\hspace*{1em}\textbullet\ May need to combine 2--3 concepts or rules\\
\hspace*{1em}\textbullet\ Example: ``How would inflation affect consumer spending patterns?''\\[4pt]

\textnormal{\textbf{4 - Complex:}} Multi-step analysis with dependencies\\
\hspace*{1em}\textbullet\ Requires 5--8 logical steps with intermediate conclusions\\
\hspace*{1em}\textbullet\ Need to track multiple variables or relationships\\
\hspace*{1em}\textbullet\ May require considering trade-offs or constraints\\
\hspace*{1em}\textbullet\ Example: ``Design an algorithm to optimize traffic flow in a city''\\[4pt]

\textnormal{\textbf{5 - Deep:}} Extensive reasoning with abstract thinking\\
\hspace*{1em}\textbullet\ Requires 8+ logical steps with nested reasoning\\
\hspace*{1em}\textbullet\ Must synthesize multiple frameworks or domains\\
\hspace*{1em}\textbullet\ Involves hypothesis formation, proof, or complex problem decomposition\\
\hspace*{1em}\textbullet\ Example: ``Prove the halting problem is undecidable''\\[6pt]

\textnormal{\textbf{Consider:}}\\
\hspace*{1em}- Number of logical steps required\\
\hspace*{1em}- Depth of decomposition needed\\
\hspace*{1em}- Dependencies between reasoning steps\\
\hspace*{1em}- Need for intermediate conclusions\\[6pt]

\textnormal{\textbf{Output format:}}\\
Analysis: [Analyze the question step by step, identifying: (1) what knowledge/operations are needed, (2) how many logical steps are involved, (3) what dependencies exist between steps, (4) which scale level best matches these requirements]\\
ReasoningRequired: [1--5]
\end{tcolorbox}
\caption{Prompt template for rating the reasoning complexity of questions. The \texttt{\{question\}} placeholder is replaced with the target question at inference time.}
\label{tab:reasoning_complexity_prompt}
\end{table*}

\end{document}